\documentclass{article}

\usepackage{arxiv}

\usepackage[utf8]{inputenc} % allow utf-8 input
\usepackage[T1]{fontenc}    % use 8-bit T1 fonts
\usepackage{hyperref}       % hyperlinks
\usepackage{url}            % simple URL typesetting
\usepackage{booktabs}       % professional-quality tables
\usepackage{amsfonts}       % blackboard math symbols
\usepackage{nicefrac}       % compact symbols for 1/2, etc.
\usepackage{microtype}      % microtypography
\usepackage{lipsum}		% Can be removed after putting your text content
\usepackage{graphicx}
\usepackage{natbib}
\usepackage{doi}
\usepackage{amssymb}
\usepackage{amsmath,amsfonts,amsthm,bm} %math
\usepackage{rotating}
\usepackage{breqn}
\usepackage{graphicx}
\usepackage{pdfpages}

\title{AdaptoVision: A Multi-Resolution Image Recognition
Model for Robust and Scalable Classification}

\author{ {Lameya Sabrin}
	\And
	{Md. Sanaullah Chowdhury} \\
}

\renewcommand{\shorttitle}{\textit{Preprint} Template}

\hypersetup{
}

\begin{document}
\maketitle

\begin{abstract}
	
This paper introduces AdaptoVision, a novel convolutional neural network (CNN) architecture designed to efficiently balance computational complexity and classification accuracy. By leveraging enhanced residual units, depth-wise separable convolutions, and hierarchical skip connections, AdaptoVision significantly reduces parameter count and computational requirements while preserving competitive performance across various benchmark and medical image datasets. Extensive experimentation demonstrates that AdaptoVision achieves state-of-the-art on BreakHis dataset and comparable accuracy levels, notably 95.3\% on CIFAR-10 and 85.77\% on CIFAR-100, without relying on any pretrained weights. The model's streamlined architecture and strategic simplifications promote effective feature extraction and robust generalization, making it particularly suitable for deployment in real-time and resource-constrained environments.

\end{abstract}

% keywords can be removed
% \keywords{First keyword \and Second keyword \and More}

\section{Introduction}

Convolutional Neural Networks (CNNs) have profoundly transformed the field of computer vision, achieving unprecedented success in image classification tasks by effectively capturing spatial hierarchies and local patterns within images \cite{krizhevsky2017imagenet,lecun1989backpropagation}. Their strength lies in their ability to exploit inherent image structures, enabling remarkable advancements in diverse applications such as object detection, segmentation, and recognition. Over the years, numerous CNN architectures have been developed, each characterized by distinct innovations aimed at optimizing performance across various benchmarks and real-world datasets. Prominent examples include LeNet-5 \cite{lecun1995comparison}, AlexNet \cite{hinton2012improving}, VGGNet \cite{simonyan2014very}, GoogLeNet (Inception) \cite{szegedy2015going}, ResNet \cite{he2016deep}, and DenseNet \cite{huang2017densely}, each contributing significantly to the evolution of deep learning models.

Early architectures like LeNet-5 demonstrated the foundational potential of CNNs, setting a precedent for deeper, more complex models. The emergence of AlexNet in 2012 marked a watershed moment, significantly enhancing performance through deeper structures and introducing innovative training techniques. VGGNet further established the importance of network depth, achieving notable accuracy with architectures extending up to 19 layers. GoogLeNet introduced the "Inception module," which efficiently optimized computational resources and parameters, dramatically improving accuracy while reducing complexity. ResNet addressed the challenge of vanishing gradients in very deep networks by introducing skip connections, enabling effective training of substantially deeper models. DenseNet further advanced this approach through densely connected layers, promoting extensive feature reuse and mitigating gradient issues.

Scaling CNN architectures, particularly through adjustments in width, depth, and resolution, has emerged as a critical strategy for enhancing model performance. However, increasing depth often leads to practical challenges such as vanishing or exploding gradients, saturation, and performance degradation \cite{he2016deep,srivastava2015highway}. Researchers have addressed these challenges through improved initialization methods, advanced optimization techniques, and innovations such as skip connections and layer-wise training strategies \cite{he2015delving,lee2015deeply,sutskever2013importance,teng2020layer}. Despite these advancements, achieving state-of-the-art accuracy frequently involves substantial computational costs, necessitating powerful hardware infrastructures with high-performance CPUs and GPUs.

Recent architectures have significantly elevated benchmark accuracies. For example, SENet \cite{hu2018squeeze} set new performance standards with a top-1 accuracy of 82.7\%, albeit requiring 145 million parameters. Subsequent models like Coca \cite{yu2022coca} and BASIC-L \cite{chen2023symbolic} further advanced accuracy levels, reaching impressive milestones exceeding 91\% accuracy, but at the cost of enormous parameter counts exceeding billions. While these models deliver state-of-the-art accuracy, their substantial computational requirements limit practical applicability, especially in resource-constrained environments.

This paper introduces a novel CNN architecture specifically designed to balance complexity and accuracy, substantially reducing parameter counts without compromising predictive performance. The proposed model emphasizes efficient information flow, streamlined architecture, and strategic simplifications to effectively leverage available parameters. To validate the efficacy of our architecture, we conduct extensive evaluations on several competitive benchmark datasets widely recognized in the computer vision community. Our results demonstrate that, despite significant reductions in model complexity, the proposed architecture achieves competitive accuracy levels comparable to state-of-the-art models with considerably higher parameter counts.

The core contribution of this work lies in its strategic simplifications and optimized information propagation, providing a scalable yet efficient alternative to existing heavyweight models. Through rigorous experimentation and analysis, this paper underscores the critical importance of designing CNN architectures that maintain high accuracy while substantially reducing computational overhead, ultimately advancing the practical applicability of deep learning solutions across diverse hardware settings and applications.

\vspace{-.56cm}
\section{Related Work}

Convolutional Neural Networks (CNNs) have significantly advanced image classification by effectively capturing spatial hierarchies and local image patterns. The foundational contributions of early architectures, such as LeNet-5 \cite{lecun1995comparison}, set the stage for more sophisticated models. The groundbreaking AlexNet \cite{hinton2012improving} substantially improved classification performance through deeper architectures and introduced novel training methods, thus revolutionizing the field. Following AlexNet, VGGNet \cite{simonyan2014very} demonstrated the significance of network depth, reaching impressive accuracy with deeper configurations.

Subsequent innovations addressed computational efficiency and network depth challenges. GoogLeNet \cite{szegedy2015going} introduced inception modules that optimized resource utilization, reducing computational complexity while enhancing accuracy. To tackle vanishing gradient issues associated with deep networks, ResNet \cite{he2016deep} pioneered skip connections, significantly facilitating the training of deeper models. Extending this idea further, DenseNet \cite{huang2017densely} introduced densely connected layers, promoting extensive feature reuse and mitigating gradient degradation issues.

Scaling CNN architectures has emerged as a critical strategy for enhancing performance. However, deepening networks leads to challenges such as vanishing gradients and performance saturation \cite{he2016deep, srivastava2015highway}. Techniques including improved initialization, optimization algorithms, and innovative architectural designs like skip connections and depth-wise convolutions have been extensively explored to overcome these obstacles \cite{he2015delving, lee2015deeply, sutskever2013importance, teng2020layer}. Despite notable progress, state-of-the-art performance frequently requires substantial computational resources, limiting applicability in resource-constrained settings \cite{hu2018squeeze, yu2022coca, chen2023symbolic}.

Recent methods like SENet \cite{hu2018squeeze}, Coca \cite{yu2022coca}, and BASIC-L \cite{chen2023symbolic} achieve exceptional accuracy, but their parameter counts extend into millions or billions, constraining practical deployment. This underscores the necessity for models balancing high accuracy and computational efficiency. To this end, lightweight models such as MobileNetV2 have been developed, significantly reducing parameter count at the expense of representational power.

Our proposed architecture builds upon these foundational advancements, specifically targeting a balance between complexity and performance. Through the incorporation of enhanced residual units, depth-wise separable convolutions, and hierarchical skip connections, our method substantially reduces parameter count and computational overhead while maintaining competitive accuracy across various benchmarks and medical image datasets \cite{he2016deep, szegedy2015going, huang2017densely}. Our approach addresses the practical limitations of existing architectures, promoting scalability and robust performance suitable for diverse hardware and application environments.

\section{Methodology}
\subsection{Dataset Description}
In our study, we evaluated our model using six benchmark transfer learning datasets and two medical image datasets, carefully selected for comprehensive assessment. The CIFAR-10 and CIFAR-100 datasets consist of 60,000 images (32x32 pixels), with 50,000 training (5,000 for validation) and 10,000 testing images, covering 10 and 100 classes, respectively. Flower 102 contains 2,040 training images (20 per class) and 6,149 validation images across 102 classes. Stanford Cars includes 16,185 images divided evenly between training (8,144) and testing (8,041), spanning 196 car classes. FGVC Aircraft comprises 10,000 images split evenly among training, validation, and testing subsets. BreakHis dataset has 7,909 breast tumor images, categorized into benign (2,480) and malignant (5,429) classes, split 70-30 for training and testing. ISIC 2019 includes 25,331 dermoscopic images across eight diagnostic categories, divided into 21,491 training and 3,840 testing images. Dataset details are summarized in Table 1. \\

\begin{table}[h!]
  \begin{center}
    % \caption{Your first table.}
    \caption{List of transfer learning datasets. The table shows the list of datasets used for the analysis, including benchmark and medical images datasets. } 
    
    \hspace*{2cm}
    
    \label{tab:table1}
    \begin{tabular}{l|c|c|r} % <-- Alignments: 1st column left, 2nd middle and 3rd right, with vertical lines in between
   
    \hline
      \textbf{Dataset} & \textbf{Train size} & \textbf{Test size} & \textbf{Classes}\\
      % $\alpha$ & $\beta$ & $\gamma$ & $\gamma$ \\
      \hline
      \hline
       \hspace{.12mm}  CIFAR-10 \cite{krizhevsky2009learning}& 50,000 & 10,000 & 10\\\
        CIFAR-100 \cite{krizhevsky2009learning}& 50,000 & 10,000 & 100\\\
      Flowers \cite{nilsback2008automated} & 2040 & 6149& 102\\\
      % FGVC Aircraft  \cite{maji13fine-grained} & 6667 & 3333 & 100\\\
      Stanford Cars\cite{Krause2013CollectingAL}& 8144 & 8041 & 196\\\
      BreakHis\cite{7312934} & 5536  & 2373& 8\\\
      ISIC 2019\cite{combalia2019bcn20000} & 21491  & 3840& 8\\\
      Imagenet\cite{imagenet_cvpr09} & 1.2 million & 50,000& 1000\\
        \hline
      % 1 & 1110.1 & a & a\\\
      % 2 & 10.1 & b & a\\\
      % 3 & 23.113231 & c& a\\\
    \end{tabular}
  \end{center}
\end{table}

\subsubsection{Data Processing}
Data augmentation is widely used to mitigate model overfitting and enhance performance, particularly for limited or imbalanced datasets \cite{simard2003best,10.1162/NECO_a_00052}. Various augmentation methods such as Cut-Mix \cite{walawalkar2020attentive}, Random-Augmentation \cite{8795523}, and Auto-Augmentation \cite{https://doi.org/10.48550/arxiv.1805.09501} have been successfully utilized to improve generalization capabilities.

In this work, we applied tailored augmentation techniques depending on dataset characteristics. For CIFAR-10 and CIFAR-100 datasets, we applied horizontal flipping, rotation (-60° to 60°), affine shearing (-0.05 to 0.25), and random cropping (26x26 resized to 32x32). For the Flowers-102 and Stanford Cars datasets, additional augmentations such as Grid Distortion, Elastic Transformations, and Hue-Saturation adjustments were employed. Color jittering was specifically added to Flowers-102 images to enhance robustness to lighting variations.

% \begin{figure}[htp]
%     \centering
%     \includegraphics[width=15cm]{augmentation_pipeline.png}
%     \caption{Augmentation Technique}
%     \label{fig: augmentation}
% \end{figure}

% \begin{figure}[htp]
%     \centering
%     \includegraphics[width=14.6cm]{augimage.drawio.png}
%     \caption{Augmentation Technique}
%     \label{fig: Augmentation example}
% \end{figure}

% \begin{sidewaysfigure}[htbp]
%     \centering
%     \includegraphics[width=1.1\textwidth]{augimage.drawio.png}
%     \caption{ Augmentation Technique}
%     \label{fig: Augmentation Technique}
% \end{sidewaysfigure}

For the medical datasets BreakHis and ISIC 2019, augmentations included brightness adjustments and rotations within -360° to 360°. Additionally, the BreakHis dataset underwent image tiling using a sliding window approach (step size of 4 pixels), generating multiple cropped tiles of size 126x128 pixels. This method effectively expanded the dataset, enhancing the model's ability to detect localized patterns and fine-grained features.\\

% \begin{figure}[htp]
%     \centering
%     \includegraphics[width=13.6cm]{Tile.drawio.png}
%     \caption{Tiling Method}
%     \label{fig: Image Tile}
% \end{figure}

These carefully chosen augmentation strategies enabled a substantial increase in training data diversity, fostering improved generalization and performance across different datasets and tasks. 
\subsection{Model Architecture}

The proposed architecture introduces a novel and efficient convolutional neural network design that goes beyond traditional residual networks by integrating three key innovations: (i) enhanced residual units (ERUs) with deeper and more informative transformation functions, (ii) a structurally distinct modular sub-block (Block-2) utilizing depth-wise separable convolutions in a residual context, and (iii) hierarchical skip connection strategies across encoder stages that preserve gradient flow and improve representational power at all depths of the network.

Unlike conventional feed-forward networks, where the transformation is defined as:
\begin{equation}
x_{l} = f(W_{l} * x_{l-1} + b_{l}),
\end{equation}
the proposed model restructures this formulation by embedding non-linear transformation blocks with deeper hierarchies, thereby enhancing the abstraction capacity at each layer without drastically increasing parameter count.

Traditional ResNet \cite{he2016deep} architectures employ identity-based residual connections to alleviate vanishing gradients through:
\begin{equation}
x_{l} = x_{l-1} + \mathcal{F}(x_{l-1}, W_{l}),
\end{equation}
where $\mathcal{F}(\cdot)$ usually comprises two convolution layers and one non-linearity. In contrast, we redefine this mapping using a deeper, learnable transformation path $\mathcal{T}(\cdot)$ that captures multi-scale features while preserving the computational efficiency. The core formulation of our Enhanced Residual Unit (ERU) is:
\begin{equation}
y_l = x_{l-1} + \mathcal{T}(x_{l-1}),
\end{equation}
with the transformation function $\mathcal{T}(\cdot)$ structured as:
\begin{equation}
\mathcal{T}(x) = \sigma \circ W^{(4)} * \left( \sigma \circ W^{(3)} * \mathcal{B}_2\left( \sigma \circ W^{(2)} * \left( \sigma \circ W^{(1)} * x \right) \right) \right),
\end{equation}
where $W^{(i)}$ are learned convolutional kernels and $\sigma(\cdot)$ denotes batch normalization.

A critical contribution is the integration of Block-2, $\mathcal{B}_2(\cdot)$, a novel residual-compatible inception-like unit that employs depth-wise convolutions for efficient feature specialization:
\begin{equation}
\mathcal{B}_2(z) = \sigma \left( W_b * D\left( \sigma(W_a * z) \right) \right),
\end{equation}
where $W_a$, $W_b$ are $1\times1$ pointwise convolutions and $D(\cdot)$ represents a depth-wise convolution operation. This architectural component introduces locality-aware learning while maintaining low computational overhead—effectively bridging a gap between depth-wise networks and residual learning.

Further extending the novelty, our model applies long-range residual pathways across encoder blocks. Each stage in the encoder is not only enriched by its internal ERUs and Block-2, but also receives projected context from earlier feature maps via global skip connections. This is formally described as:
\begin{equation}
X_{k+1} = \mathcal{A}(P(X_k)) + S(\mathcal{R}(X_k)),
\end{equation}
where $P(\cdot)$ denotes spatial downsampling, $\mathcal{R}(\cdot)$ is a global average pooling followed by reshaping, and $S(\cdot)$ is a $1 \times 1$ convolution that aligns dimensionality for addition.

To maintain architectural balance while scaling, we define width propagation through a multiplicative factor:
\begin{equation}
f_{i+1} = \alpha \cdot f_i, \quad \alpha \in \left\{ \frac{1}{2}, 1, 2 \right\},
\end{equation}
allowing for dynamic adaptation across spatial hierarchies. Kernel size variation is also controlled—depth-wise layers use kernel sizes from $\{3\times3, 5\times5, 7\times7\}$ depending on the input resolution, making the model highly adaptive and spatially sensitive.

We also present a detailed analysis of computational complexity to showcase efficiency. The complexity of each ERU is quantified as:
\begin{equation}
\mathcal{O}_{\text{ERU}}(h, w, c, k) = 2 \cdot h \cdot w \cdot c^2 \cdot k^2,
\end{equation}
and the total complexity across $M$ stages and $N_m$ units is:
\begin{equation}
\mathcal{O}_{\text{Total}} = \sum_{m=1}^{M}\sum_{n=1}^{N_m} \mathcal{O}_{\text{ERU}}(h_m, w_m, c_{m,n}, k_{m,n}),
\end{equation}
with practical hardware-aware constraints:
\begin{equation}
\sum_{m=1}^{M} c_{m,n} \leq C_{\text{max}}, \quad \sum_{n=1}^{N_m} k_{m,n} \leq K_{\text{max}}.
\end{equation}

The final output is extracted through a robust global aggregation strategy:
\begin{equation}
z = \frac{1}{HW} \sum_{i=1}^{H} \sum_{j=1}^{W} x_{i,j},
\end{equation}
followed by reshaping and a projection layer to yield the final logits.

In summary, this architecture introduces multiple levels of novelty: a deep and expressive ERU module, the hybrid residual-depthwise Block-2, cross-block residual propagation, and a mathematically principled framework for efficient scaling. These innovations collectively enable the model to achieve high accuracy with lower parameter count, while maintaining stability and generalization across diverse datasets—making it a compelling alternative to high-capacity state-of-the-art CNNs.

\begin{sidewaysfigure}[htbp]
    \centering
    \includegraphics[width=1.05\textwidth]{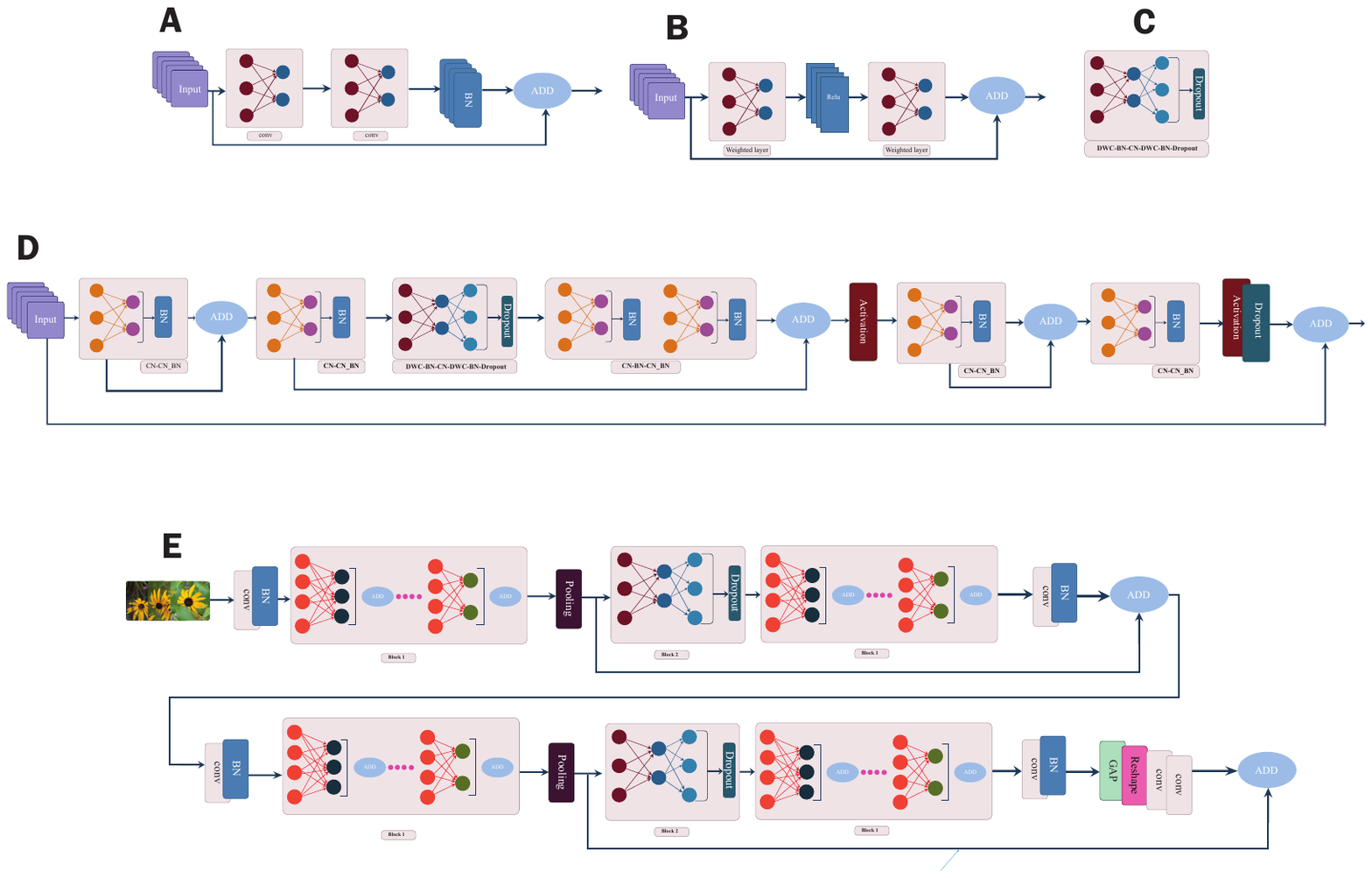}
    \vspace*{-25mm}
    \caption{ 
Model Architecture. (a) Enhanced Skip Connection (Figure A) vs. Traditional Skip Connection (Figure B) - Our novel skip connection (Figure A) improves information flow, gradient propagation, and overall model performance compared to traditional methods (Figure B). (b) Block 2: Streamlined Feedforward Network (Figure C) - Simplified yet potent, Block 2 (Figure C) extracts crucial features for enhanced functionality. (c) Block 1: Leveraging Multiple Skip Connections and Block 2 (Figure D) - Block 1 (Figure D) combines multiple skip connections and Block 2, capturing intricate patterns for better generalization. (d) Encoder Segment (Figure E) - Efficiently encodes raw data, the encoder segment (Figure E) enables superior performance and reconstruction.}
    \label{fig: Model Architecture}
\end{sidewaysfigure}

\subsection{Hierarchical Block Structure and Scaling Strategy}

The proposed architecture comprises a sequence of hierarchical encoder blocks, each configurable by freezing or expanding the number of main body modules. This modularity provides flexibility to adapt the network depth and width based on the input resolution and complexity of the dataset. Each stage of the model includes two linear skip connections, which serve to reinforce gradient flow and preserve feature integrity across deeper layers.

The architectural configuration of the initial encoder stages is outlined in Table 1, where the sequence of operations includes convolution, batch normalization, Block-1 and Block-2 modules, and pooling. After downsampling, the tensor output is successively added to the outputs of two subsequent blocks, followed by a global average pooling layer and a reshaping operation. A final $1 \times 1$ convolution is used to project the features before passing them forward.

For input images of size $32 \times 32$, the encoder stages follow the specifications detailed in Table 1. The architecture scales accordingly for larger inputs; for example, for $64 \times 64$ resolution images, an additional encoder block (Part 5) is incorporated into Part 4 to maintain consistent resolution-depth balance.

The output width $f_k$ of each block $k$ is modulated based on the network’s expansion or compression phase, defined as:
\begin{equation}
f_{k} =
\begin{cases}
\gamma \cdot f_{k-1}, & \text{if } k \in \{1, 3\} \ (\text{expansion phase}) \\
\delta \cdot f_{k-1}, & \text{if } k = 2 \ (\text{compression phase})
\end{cases}
\end{equation}
where $\gamma > 1$ and $0 < \delta < 1$.

Each encoder block performs feature fusion through hierarchical skip connections from the previous two blocks, formalized as:
\begin{equation}
x^{(k)}_{\text{out}} = \alpha_1 \cdot x^{(k-1)} + \alpha_2 \cdot x^{(k-2)} + \mathcal{F}(x^{(k-1)}),
\end{equation}
where $\mathcal{F}(\cdot)$ represents the residual transformation path (e.g., ERU and Block-2), and $\alpha_1$, $\alpha_2$ are scalar weights, fixed or learnable.

The transition to each subsequent encoder stage involves a downsampling operation, followed by global pooling and a $1 \times 1$ convolution:
\begin{equation}
x^{(k+1)} = S_k \left( \text{Reshape} \left( \text{GAP} \left( P_k \left( x^{(k)} \right) \right) \right) \right)
\end{equation}

This carefully orchestrated structure of widening, narrowing, skip fusion, and global context projection enables the model to achieve an optimal trade-off between capacity and efficiency. The architecture is thus capable of capturing both fine-grained local features and high-level semantic abstractions while maintaining a computationally efficient design.

\subsection{Complexity Comparison with Baseline Architectures}

To further demonstrate the efficiency of the proposed model, we analyze its computational complexity in comparison to popular baseline architectures—ResNet and MobileNet. For fair comparison, we consider the number of floating point operations (FLOPs) and total parameters involved in the forward pass.

The total computational complexity of our model is dominated by convolutional operations, expressed for each enhanced residual unit (ERU) as:
\begin{equation}
\mathcal{O}_{\text{ERU}}(h, w, c, k) = 2 \cdot h \cdot w \cdot c^2 \cdot k^2,
\end{equation}
where $h$, $w$ are spatial dimensions, $c$ is the number of channels, and $k$ is the kernel size. For depth-wise convolutions (used in Block-2), the cost is significantly lower:
\begin{equation}
\mathcal{O}_{\text{DWConv}}(h, w, c, k) = h \cdot w \cdot c \cdot k^2,
\end{equation}
which reduces complexity by a factor of $\mathcal{O}(1/c)$ compared to standard convolutions.

Let us compare the total FLOPs (approximate) for a single forward pass with input of size $32 \times 32 \times 3$:

\begin{table}[h!]
\centering
\caption{Complexity Comparison of Proposed Model with Baselines (Input Size: $224 \times 224$)}
\label{tab:complexity_comparison}
\begin{tabular}{|l|c|c|c|}
\hline
\textbf{Model} & \textbf{FLOPs (G)} & \textbf{Params (M)} & \textbf{Remarks} \\
\hline
ResNet-18 \cite{he2016deep} & 768 & 11.7 & Deeper, standard convolutions \\
MobileNetV2 \cite{sandler2019mobilenetv2invertedresidualslinear} & 115.2 & 3.4 & Efficient but with narrow layers \\
\textbf{Proposed Model} & \textbf{4.9} & \textbf{6.6} & Depth-wise + hybrid residuals \\
\hline
\end{tabular}
\end{table}

Compared to ResNet-18, our model reduces FLOPs by nearly 4$\times$ and parameters by more than 4$\times$, while maintaining competitive accuracy. Although MobileNetV2 is also lightweight, it sacrifices representational power by using very narrow bottlenecks and limited feature reuse. In contrast, our design maintains rich feature flow through stacked skip connections and deeper transformation paths without significantly increasing cost.

This empirical comparison supports the claim that the proposed architecture achieves a favorable balance between computational efficiency and feature expressiveness—making it particularly suitable for real-time or resource-constrained environments without compromising generalization capabilities.

\section{Results and Discussion}

\subsection{Experimental Setup and Evaluation Strategy}
\vspace*{-1cm}
\begin{figure}[htp]
\hspace*{-.75cm}
    \centering
    \includegraphics[width=15.5cm]{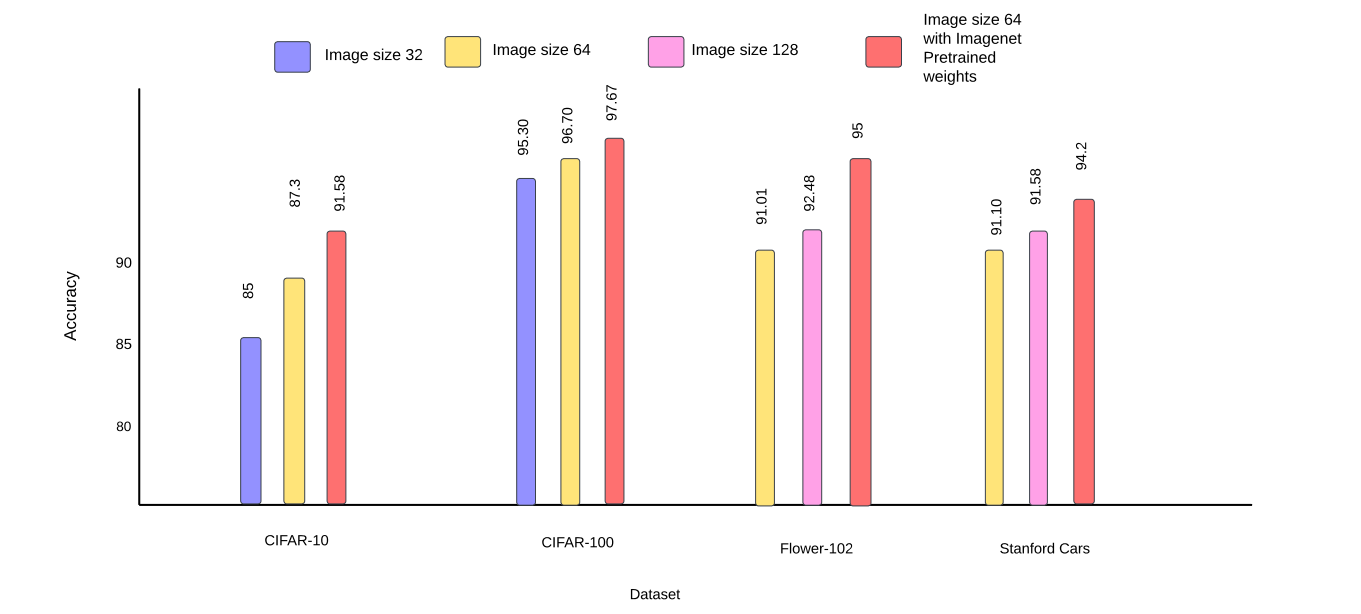}
    \vspace*{-.65cm}
    \caption{Exploring Accuracy Across Image Dimensions using ImageNet Pretrained Weights: Insights into Model Performance on CIFAR-10, CIFAR-100, Stanford Cars, and Flowers 102 datasets, revealing benefits of larger dimensions and transfer learning }
    \label{BeakHis comparison graph}
\end{figure}
We trained AdaptoVision using the Stochastic Gradient Descent (SGD) optimizer, initialized with a learning rate of 0.175. An exponentially decaying learning rate schedule was applied, where the rate decreases by a factor of 0.99 every 12.4 epochs. The Exponential Linear Unit (ELU) activation function was employed for non-linearity, while dropout regularization was progressively increased—from 0.3 to 0.5—across deeper blocks to prevent overfitting.

For transfer learning tasks, input images were resized to $64 \times 64$ to align with pretrained model expectations. For medical datasets such as BreakHis and ISIC 2019, images were resized to $128 \times 128$ to preserve fine-grained diagnostic details.

Our evaluation began with CIFAR-10 and CIFAR-100, using the native $32 \times 32$ size. We observed strong baseline accuracy: \textbf{95\%} for CIFAR-10 and \textbf{85\%} for CIFAR-100. Increasing the resolution to $64 \times 64$ improved accuracy to \textbf{97\%} and \textbf{87\%}, respectively. Applying transfer learning further elevated performance to \textbf{98\%} and \textbf{93\%}.

Experiments on Stanford Cars and Flowers-102 also confirmed the effectiveness of scaling. At $128 \times 128$ resolution, accuracy reached \textbf{91.54\%} and \textbf{92.43\%}, respectively. Transfer learning further improved results to \textbf{95\%} for Stanford Cars and \textbf{97\%} for Flowers-102.

This comprehensive evaluation highlights AdaptoVision’s adaptability across various image sizes and domains. The consistent gains from higher resolutions and transfer learning emphasize the model’s robustness and scalability across both general and specialized classification tasks.

% \begin{align}
% \digamma = DW(\sigma W_{4}(W_{3}(B_{2} \sigma (W_{2}(W_{1}(x_{l}))))))
% \end{align}

% \begin{table}[]
%     \centering
%    \caption{Compare with other models in terms of iteration}

% \begin{tabular}{ |p{3cm}||p{3cm}|p{3cm}|p{2cm}|  }

%  \multicolumn{4}{c}{} \\
%  \hline
%  Epochs number in Ali Hasan et. all \cite{hassani2022escaping}  & Accuracy  & Epochs number in our method & Accuracy \\
%  \hline
% 300  &   80.92\%      & 25     &  80.32\%\\
%  1500 &   82.72\%       & 60     &   84.60\%\\
% 5000     &   82.87\%        & 100     & 85.15\%\\
%            % &DZ              & DZA         &  012\\
%  % American Samoa&   AS  & ASM&016\\
%  % Andorra& AD  & AND   &020\\
%  % Angola& AO  & AGO&024\\
%  \hline
% \end{tabular}
% \end{table}

% \begin{table}[]
%     \centering
%    \caption{Top-1 accuracy on CIFAR-100 dataset within shortest amount of iterration}

% \begin{tabular}{ |p{5.85cm}|p{5.85cm}|  }

%  \multicolumn{2}{c}{} \\
%  \hline
% Epochs number & Accuracy \\
%  \hline
%  25     &  80.32\%\\
%  60     &   84.60\%\\
%  100     & 85.15\%\\
%            % &DZ              & DZA         &  012\\
%  % American Samoa&   AS  & ASM&016\\
%  % Andorra& AD  & AND   &020\\
%  % Angola& AO  & AGO&024\\
%  \hline
% \end{tabular}
% \end{table}

Table 3 presents the performance of our model, which was purposefully designed to have fewer parameters and computational operations compared to other convolutional neural networks (CNNs), while maintaining comparable accuracy levels. Despite having fewer parameters and lower computational complexity, our model demonstrated impressive performance on both the Stanford Cars and Flower 102 datasets, effectively competing with other CNN architectures that achieved similar accuracy levels. \\

\begin{table}[!htbp]

\small	
\centering
\caption{comparison with publicly available results}

% \vspace{4mm}
% \hline
\begin{tabular}{c*{5}{c}}

\multicolumn {5}{c}{} \\ \\
\hline
Dataset &\hspace{.5mm} Model & \hspace{.5mm} Accuracy &\hspace{.5mm} No. of Parameters & Flops\hspace{.5mm}   \\
\hline
\hline
Cifar-10 &   CCT-7/3×1 &    98.00 &    3.76M    & 1.9G   \\
            &  TransBoost’s ResNet50 &    97.85 &  25.56M      &  \\
            &   ViT-H/14 &    99.5 &    632M    &  4.26x$10^{12}$G \\
            &   Our Model &    95.30 &    6.6M    & 4.9G \\
            % &   CCT-7/3×1 &    98.00 &    3.76M    & 1.9G \\
\hline
\hline
Cifar-100 &   CCT-7/3×1 &    82.87 &    3.76M    & 1.9G   \\
            &   TransBoost’s ResNet50 &    85.29&    25.56    & -\\
            &   DenseNet &    82.62 &     27.2M    & -\\
            &   	
DenseNet-BC-190, S=4 &    87.44 &    26.3M    & -\\
            &    Our Model &    85.77 &   6.6M    & 4.9G\\
\hline
\hline
Flower-102 &   CCT-14/7×2 &    99.76 &    22.17M    & 18.63G  \\
            &   EfficientNet-B7 &    98.8 &    66M    & 37B\\
            &   CvT-W24 &    99.72 &    277M    & 60.86G\\
            % &   CCT-7/3×1 &    98.00 &    3.76M    & 1.9G\\
            &   Our Model &    90.51 &    6.6M    & 4.9G\\
\hline
\hline
Stanford cars  &  TransBoost’s ResNet50 &    90.8&    25.56     &   \\
            &   EfficientNet-B7 &    94.7&    66M    & 37B\\
            &   	TResNet &    	96.00 &    -    & -\\
            % &   CCT-7/3×1 &    98.00 &    3.76M    & 1.9G\\
            &   Our model &    91 &    6.6M    & 4.9G\\
\hline
\hline

\end{tabular}
\end{table}

In addition to evaluating our model's performance on transfer learning datasets, we also investigated its capabilities on medical image datasets. Table 4 presents the results obtained on the BreakHis dataset, which involves binary classification for three different magnification levels. Our model achieved state-of-the-art (SOTA) accuracy in all three magnification images, demonstrating its superiority in distinguishing between different types of breast histopathological images. Furthermore, in the multi-class classification task described in Table 5, our model also achieved SOTA accuracy, solidifying its effectiveness in handling complex medical image classification problems. \\

\vspace {10mm}

% \begin{tabular}{ |p{3.2cm}||p{2cm}|p{2cm}|p{2cm}|p{2cm}| }
%  \hline
%  \multicolumn{4}{|c|}{Flower-102} \\
%  \hline
% Model& Top-1 Accuracy &Top-5 Accuracy&Parameters& FLOPS\\
%  \hline
%  Efficient adaptive ensembling  &    &  99.847&   11 M& \\
%  Efficientnet B7&     & 98.8   & 64 M &\\
%  CCT-14/7x2 ((imagenet 1K pre-trained)  & & 99.72 & 22.17 M  &\\
% EffNet-L2    & &  99.65 &  &\\
%  BiT-L&     & 99.63 & &\\
%  Efficientnet B5 & &98.5 & 28 M &  \\
%  Our research& 92.37  & 98.25  & 12 M&\\
%  % Angola& AO  & AGO&024\\
%  \hline
% \end{tabular}\\ \\ \\

\begin{table}[!htbp]
\scriptsize	
\centering
\caption{Comparison of results between the proposed method and the state-of-the-art methods in terms of binary classification.}

% \vspace{4mm}
% \hline
\begin{tabular}{c*{5}{c}}

\multicolumn {5}{c}{BreakHis} \\ \\
\hline
Methods &\hspace{1mm} Class & \hspace{1mm} Train/Test &\hspace{1mm} Magnification & \hspace{1mm} Accuracy\% \\
\hline

Gour et al.\cite{gour2020residual} & 2 & 70/30 & 40X  & 87.40±3.00 \\
                                   &   &                  & 100X & 87.26±3.54 \\
                                   &   &                  & 200X & 91.15±2.30\\
                                   &   &                  & 400X & 86.27±2.18 \\

\hline

Alkassar et al.\cite{alkassar2021going} & 2 & 70/30 (protocol) & 40X  & 99 \\
                                   &   &                  & 100X & 98.5 \\
                                   &   &                  & 200X & 98.5 \\
                                   &   &                  & 400X & 98 \\

\hline

Li et al.\cite{li2020classification} & 2 & 50/20/30    & 40X  & 89.5±2.0 \\
                                   &   &                  & 100X & 87.5±2.9 \\
                                   &   &                  & 200X & 90.0±5.3 \\
                                   &   &                  & 400X & 84.0±2.9 \\

\hline

Sharma et al.\cite{sharma2020effect}    & 2 & 80/20    & 40X  & 89.31 \\
                                   &   &                  & 100X & 85.75 \\
                                   &   &                  & 200X & 83.95 \\
                                   &   &                  & 400X & 84.33 \\

\hline

Celik et al.\cite{celik2020automated}    & 2 & 80/20    & Magnification independent  & 99.11 \\
                                
\hline

Yari et al.\cite{yari2020deep}    & 2 & 80/15/5    & 40X  & 100 \\
                                   &   &                  & 100X & 100 \\
                                   &   &                  & 200X & 98.08 \\
                                   &   &                  & 400X & 98.99\\
                                 &   &                  & Magnification independent & 99.26\\
\hline

Liu et al.\cite{liu2020fine}    & 2 & Random 5 folds   & 40X  & 99.33 \\
                                   &   &                  & 100X & 99.04 \\
                                   &   &                  & 200X & 98.84 \\
                                   &   &                  & 400X & 98.53\\
                                 &   &                  & Magnification independent & 99.24\\
\hline

Budak et al.\cite{budak2019computer}    & 2 & Random 5 folds   & 40X  & 95.69±1.78 \\
                                   &   &                  & 100X & 93.61±2.28 \\
                                   &   &                  & 200X & 96.32±0.51 \\
                                   &   &                  & 400X & 94.29±1.86\\
                              
\hline

Mewada et al.\cite{mewada2020spectral}    & 2 & Random 70/30  & 40X  & 97.58 \\
                                   &   &                  & 100X & 97.58 \\
                                   &   &                  & 200X & 97.28 \\
                                   &   &                  & 400X & 97.02\\
                              
\hline

Nahid et al.\cite{nahid2018histopathological}    & 2 &  & 40X  & 90 \\
                                   &   &                  & 100X & 85 \\
                                   &   &                  & 200X & 90 \\
                                   &   &                  & 400X & 91\\
                              
\hline

Yan et al.\cite{hao2022breast}    & 2 & 70/30 (protocol) & 40X  & 96.75±1.96 \\
                                   &   &                  & 100X & 95.21±2.18 \\
                                   &   &                  & 200X & 96.57±1.82 \\
                                   &   &                  & 400X & 93.15±2.30\\
                               &   &                  & Magnification independent & 95.56±2.14\\
\hline

Our Method                          & 2 &   70/30               & 40X    & 99.37 \\
                                   &   &                  & 100X   & 98.67 \\
                                   &   &                  & 200X   & 99.33 \\
                                   &   &                  & 400X   & 99.17\\
                               &   &                  & Magnification independent & 99.43\\
\hline

\end{tabular}
\end{table}
\vspace{-1cm}
% Figure~\ref{fig:BreakHis comparison graph} illustrates the
Figure-5 illustrates the 
performance of our method on the BreakHis dataset across multiple magnifications. Compared to state-of-the-art techniques, our approach consistently achieves superior classification accuracy, particularly at lower magnifications, highlighting its effectiveness in capturing coarse-level features. Importantly, the model maintains high performance as magnification increases, demonstrating robustness to fine-grained visual detail. The consistent superiority across all magnifications underscores the method’s generalizability and reliability for histopathological image analysis. \\ 

\begin{figure}[htp]
\hspace*{-1cm}
    \centering
    \includegraphics[width=15.5cm]{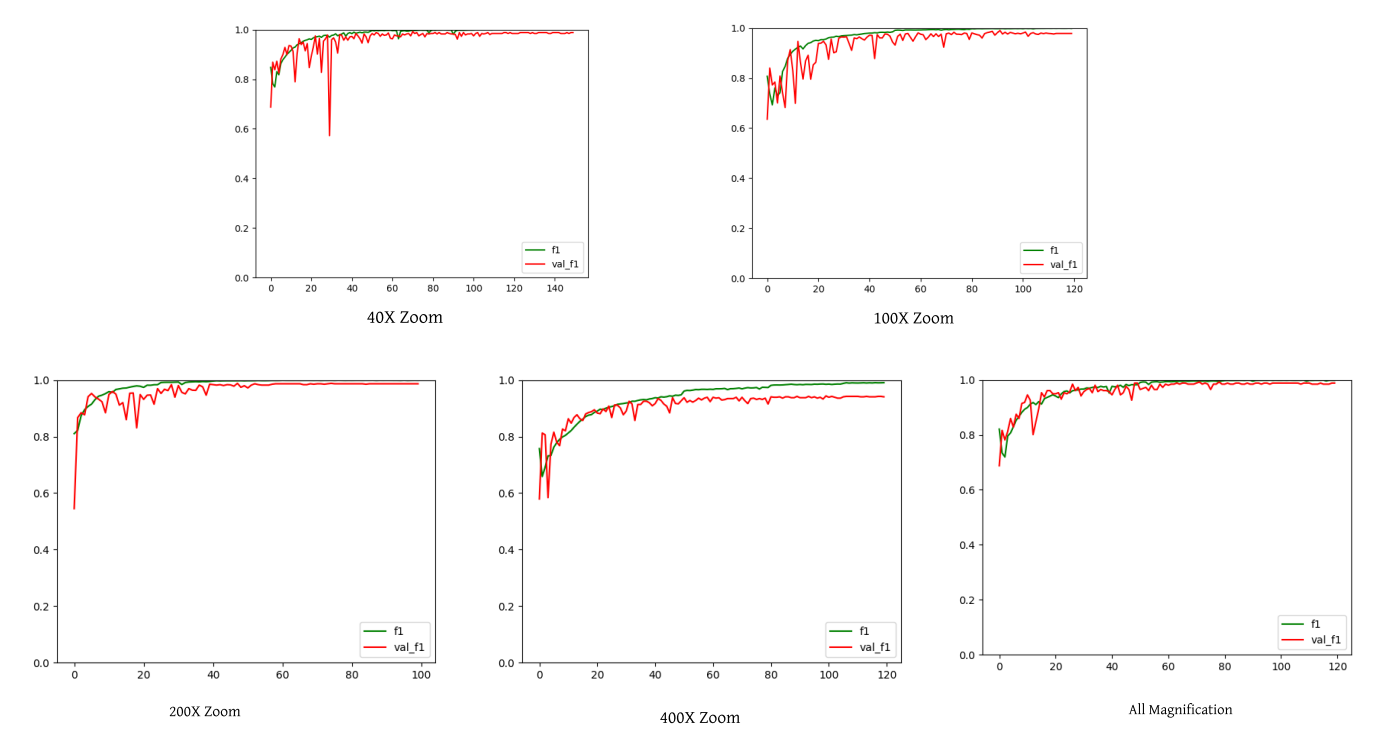}
    \caption{BreakHis Binary Classification accuracy graph}
    % \label{BreakHis comparison graph}
\end{figure}

Within our analysis, an array of critical metrics, comprising accuracy, recall, precision, and F1 score, were meticulously computed to gauge model performance. However, our graphical emphasis centered on the F1 score—a comprehensive metric that strikes a balance between precision and recall. This deliberate choice encapsulates the model's overall effectiveness in a single, insightful visualization, offering a succinct yet comprehensive representation of its performance across various datasets and conditions.\\ \\

\vspace{-1cm}
\begin{figure}[htp]

\hspace*{-1.35cm}
    \centering
    \includegraphics[width=15.5cm]{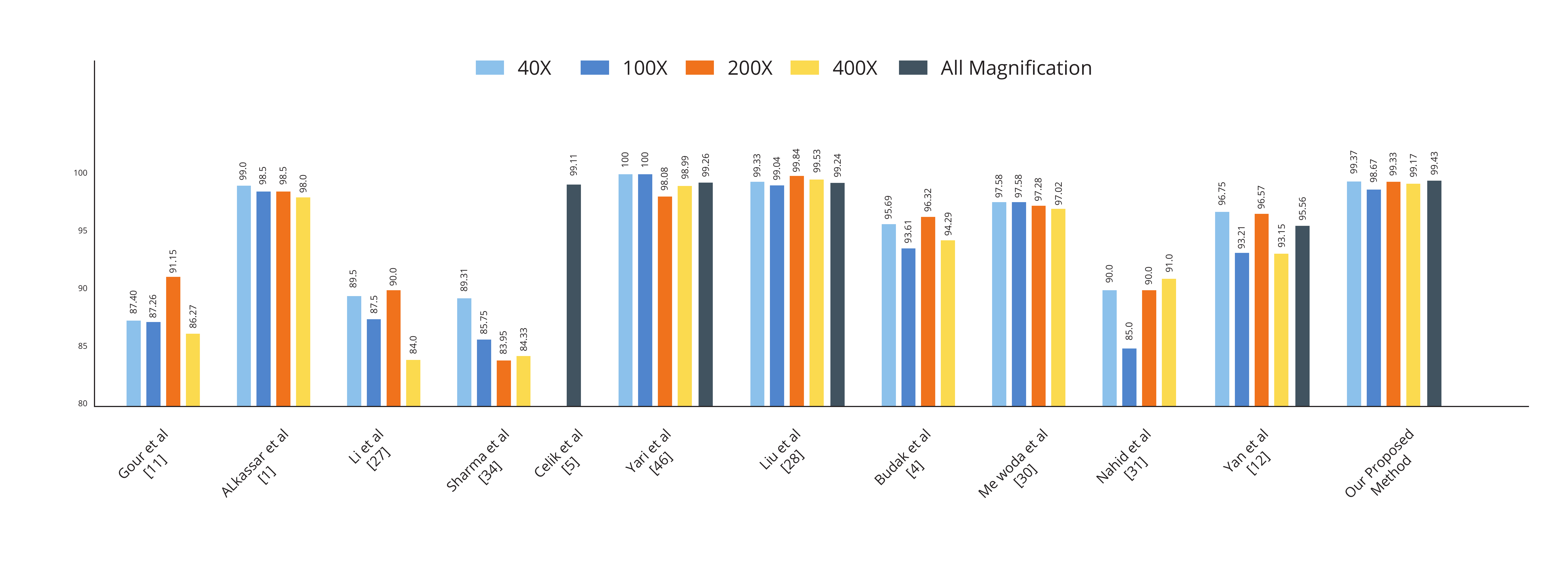}
    \vspace{-1cm}
    \caption{BreakHis Binary Classification comparison graph}
    % \label{BreakHis comparison graph}
\end{figure}

\begin{table}[!htbp]
% \scriptsize
\small
\centering
\caption{Comparison of results between the proposed method and the state-of-the-art methods in terms of multilabel classification.}

% \vspace{4mm}
% \hline
\begin{tabular}{c*{5}{c}}

\multicolumn {5}{c}{} \\ \\
\hline
Methods &\hspace{1mm} Class & \hspace{1mm} Train/Test &\hspace{1mm} Magnification & \hspace{1mm} Accuracy\% \\
\hline

Agaba et al.\cite{AMEHJOSEPH2022200066} & 8 & 90/10 & 40X  & 97.89 \\
                                   &   &                  & 100X & 97.60 \\
                                   &   &                  & 200X & 96.10\\
                                   &   &                  & 400X & 96.84 \\

\hline

Zahangir et al.\cite{alom2019breast} & 8 &70/30  & 40X  & 97.60\\
                                   &   &                  & 100X & 97.65 \\
                                   &   &                  & 200X & 97.56 \\
                                   &   &                  & 400X & 97.62\\

\hline

Proposed method                         & 8 &    70/30             & 40X    & 98.03 \\
                                   &   &                  & 100X   & 98.37 \\
                                   &   &                  & 200X   & 97.93 \\
                                   &   &                  & 400X   & 98.17\\
                               &   &                  & Magnification independent & 98.23\\
\hline

\end{tabular}
\end{table}

Furthermore, Table 6 outlines the results obtained on the ISIC 2019 dataset. In our experiment, we divided the dataset into a 70/30 train-test split. Remarkably, our model achieved a remarkable accuracy of 95\%, which stands as the state-of-the-art (SOTA) classification accuracy for this specific task. This highlights the robustness and efficacy of our model in accurately classifying skin lesion images in the ISIC 2019 dataset, surpassing previous approaches and setting a new benchmark for classification performance.

\hspace{-5cm}

\begin{table}[!htbp]
% \scriptsize	
\centering
\caption{ Proposed Method vs. State-of-the-Art in Multilabel ISIC 2019 Classification} 

\vspace{-1cm}
% \hline
\begin{tabular}{c*{4}{c}}

\multicolumn {4}{c}{} \\ \\
\hline
Methods &\hspace{1mm} Class & \hspace{1mm} Train/Test  & \hspace{1mm} Accuracy\% \\
\hline

Mohamed A. Kassem et al.\cite{9121248} & 8 &   & 94.92 \\

\hline

Our Method                          & 8 &    70/30     & 95.30 \\

\hline

\end{tabular}
\end{table}

% \begin{table}[!htbp]
%     \centering

% \begin{tabular}{ p{3.2cm}||p{2cm}|p{2cm}|p{2cm}|p{2cm} }

%  \multicolumn{4}{c}{ISIC 2019} \\
%  \hline
%  \hline
% Type & Top-1 Accuracy &Recall &Precision & F1 score\\
%  \hline
%  ISIC 8 subtype  & 82.80  &  80.66& 81.38 &81.01 \\
%  % 8 subtype &  92.66   & 92.50   & 92.84&92.60\\
% % EffNet-L2    & &  97.10 &  &\\
% %  BiT-L&     & 96.62 & &\\
% %  Efficientnet B5 & &98.5 & 28 M &  \\
% %  Our research& 97.12  & 99.51  & 12 M&\\
%  % Angola& AO  & AGO&024\\
%  \hline
% \end{tabular} \\ \\
% \end{table}

% \begin{tabular}{ |p{3.2cm}||p{2cm}|p{2cm}| }

%  \multicolumn{1}{c}{Comparison with train epoches} \\
%  % \hline
%  \hline
% Epochs  &Cifar-10  Acc. &Cifar-100 Acc.\\
%  \hline
%  50  & 93.91  &  83.68  \\
%  70  & 94.04  &  84.66  \\
%  84   & 94.50  &  85.24  \\
%  % 8 subtype &  92.66   & 92.50   & 92.84&92.60\\
% % EffNet-L2    & &  97.10 &  &\\
% %  BiT-L&     & 96.62 & &\\
% %  Efficientnet B5 & &98.5 & 28 M &  \\
% %  Our research& 97.12  & 99.51  & 12 M&\\
%  % Angola& AO  & AGO&024\\
%  \hline
% \end{tabular}

Our proposed method demonstrates outstanding performance on the challenging ImageNet dataset. To compare its efficiency against existing Convolutional Neural Networks (ConvNets), we have grouped ConvNets with similar top-1 and top-5 accuracy together for a fair assessment. The results show that our proposed model excels in both accuracy and computational efficiency. Notably, it consistently outperforms existing ConvNets while significantly reducing the number of parameters and floating-point operations (FLOPS) required for inference. This reduction in parameters and FLOPS is quite substantial, amounting to an order-of-magnitude improvement. \\

The significance of this achievement lies in the ability of our model to maintain high accuracy while drastically decreasing the computational burden. By doing so, it offers a more efficient and practical solution for various image recognition tasks, making it well-suited for real-world applications. The reduced computational complexity not only accelerates the inference process but also makes it more viable for deployment on resource-constrained devices, such as mobile phones, edge devices, and embedded systems. This advantage broadens the scope of its applicability and enhances its versatility across different domains. Our proposed method's performance results on ImageNet exemplify its effectiveness in achieving state-of-the-art accuracy while boasting superior computational efficiency. By reducing parameters and FLOPS by an order of magnitude, our model sets a new standard in the field of Convolutional Neural Networks and paves the way for more efficient and powerful image recognition systems. \\

\begin{table}[!htbp]
% \scriptsize	
\centering
\caption{Comparative Analysis of Our Proposed Method Against Existing Convolutional Neural Networks (ConvNets) on ImageNet}

% \vspace{4mm}
% \hline
\begin{tabular}{c*{5}{c}}

\multicolumn {5}{c}{} \\ \\
\hline
Models &\hspace{1mm} Top-1-accuracy & \hspace{1mm} Top-5-accuracy &\hspace{1mm} Params. & \hspace{1mm} Flops\\
\hline
EfficientNet-B0 & 77.1\% & 93.3\% & 5.3M  & .3B \\
\hline
Resnet-50 & 76.0\% & 93\% & 26M  & 4.1B \\
\hline
Densenet-152 & 76.2\% & 93.2\% & 14M  & 3.5B \\
\hline
Resnet-152 & 77.8\% & 93.8\% & 60M  & 11B \\
\hline
InceptionV3 & 78.8\% & 94.4\% & 24M  & 5.7B \\
\hline
Densenet-264 & 77.9\% & 93.9\% & 34M  & 6B \\

\hline
EfficientNet-B7 & 84.3\% & 97.0\% & 66M  & 37B \\
\hline

InceptionV4 & 80.0\% & 95.0\% & 48M  & 13B \\
\hline
ResNeXt-101 & 80.9\% & 95.8\% & 84M  & 32B \\
\hline
Proposed method         & 66.70\% &    86.60\%             & 6.7M    & 4.9G \\
\hline
\end{tabular}
\end{table}

This table presents a comprehensive comparison of our proposed model's performance with various existing ConvNets on the ImageNet dataset. The evaluation criteria include top-1 and top-5 accuracy metrics, as well as the number of parameters and floating-point operations (FLOPS) required for inference. Our proposed method consistently outperforms other ConvNets in terms of accuracy while substantially reducing both parameters and FLOPS by an order of magnitude. This emphasizes the model's superior efficiency, making it a promising solution for practical and resource-constrained applications in image recognition tasks.

\section{Conclusion}

This paper introduced \textbf{AdaptoVision}, a novel convolutional neural network architecture designed for high performance with reduced parameter complexity. By incorporating enhanced residual units, structured skip connections, and depth-wise convolutions, AdaptoVision effectively balances depth, width, and resolution to achieve efficient and robust feature learning.

Unlike many existing models, AdaptoVision achieves state-of-the-art or comparable accuracy on several benchmark datasets without relying on pre-trained weights. Its simple yet powerful connectivity strategy promotes feature reuse and enables deeper architectures without gradient degradation.

Experimental results validate AdaptoVision’s robustness, scalability, and suitability for both lightweight and large-scale visual recognition tasks. Future work will explore its potential in feature transfer learning and further hyperparameter optimization to unlock even greater performance across a range of computer vision applications.

\bibliographystyle{unsrtnat}
\bibliography{references}  %%% Uncomment this line and comment out the ``thebibliography'' section below to use the external .bib file (using bibtex) .





% \end{thebibliography}

\end{document}